\documentclass[11pt]{article} 
\usepackage[final]{acl}
\usepackage{times}

\usepackage{amsmath,amsfonts,bm}

\def\eqref#1{equation~\ref{#1}}

\def\1{\bm{1}}

\DeclareMathAlphabet{\mathsfit}{\encodingdefault}{\sfdefault}{m}{sl}
\SetMathAlphabet{\mathsfit}{bold}{\encodingdefault}{\sfdefault}{bx}{n}

\usepackage{hyperref}
\usepackage{url}
\usepackage{graphicx}
\usepackage{booktabs}

\title{Thinking in Different Spaces: \\ Domain-Specific Latent
Geometry Survives Cross-Architecture Translation}

\author{
  Marcus Armstrong \quad Navid Ayoobi \quad Arjun Mukherjee \\
  Department of Computer Science, University of Houston \\
  Houston, TX 77204 \\
  \texttt{\{miarmstr, nyoobi\}@cougarnet.uh.edu}, 
  \texttt{arjun@cs.uh.edu}
}

\begin{document}

\maketitle

\begin{abstract}
We investigate whether independently trained language models converge to 
geometrically compatible latent representations, and whether this 
compatibility can be exploited to correct model behavior at inference time 
without weight updates. We learn a linear projection matrix mapping 
activation vectors from a teacher model into a student model's coordinate 
system, then intervene on the student's residual stream during generation 
by substituting its internal state with the translated teacher 
representation. Across 20 heterogeneous teacher-student pairings, the 
projection achieves $R^2 \approx 0.50$ on verbal reasoning and 
$R^2 \approx 0.40$ on mathematical reasoning, with behavioral correction 
rates of $14.0\%$--$50.0\%$ on TruthfulQA and $8.5\%$--$43.3\%$ on GSM8K, 
demonstrating generalization across reasoning domains. Geometric alignment 
quality does not predict correction rate ($r \approx -0.07$), suggesting 
directional accuracy matters more than variance explained. Student 
sensitivity profiles are architecture-specific and invert across domains: 
the most steerable verbal student becomes the least steerable mathematical 
student. Finally, a double dissociation across all 20 pairings confirms 
that projection matrices collapse catastrophically when transferred across 
domains (mean $R^2 = -3.83$), establishing domain-specific subspace 
geometry as a property confirmed across all 20 tested pairings.
\end{abstract}

\section{Introduction}

The remarkable convergence of large language models (LLMs) on similar 
outputs suggests they might be learning the exact same underlying structure 
of language. However, it remains an open question whether their internal 
representations reflect a shared reality. Current interpretability research 
overwhelmingly focuses on intra-model dynamics---extracting, analyzing, and 
manipulating features within a single, isolated network. We extend this 
paradigm to inter-model steering. By mapping activations from one 
architecture directly to another, we probe whether independently trained 
models actually share a fundamental topological structure in their latent 
spaces.

We hypothesize that models trained on overlapping linguistic distributions 
naturally converge to topologically similar manifolds for core semantic 
concepts. If this holds true, their internal representations should not be 
entirely alien to one another. Instead, a straightforward linear 
transformation---essentially a combination of rotation and scaling---should 
be sufficient to align these manifolds. To test this, we learn a projection 
matrix that maps the intermediate activation space of a ``teacher'' model 
into the coordinate system of a ``student'' model. We then intervene on the 
student during inference, substituting its internal state with the translated 
teacher state to correct reasoning trajectories in real time.

Our findings demonstrate that this cross-architecture alignment is not only 
possible but highly effective, though it is bounded by strict geometric 
constraints. Specifically, our primary contributions are as follows:

\begin{itemize}
    \item \textbf{Empirical Existence Proof:} We show that a linear 
    projection consistently aligns latent representations across highly 
    heterogeneous architectures for both verbal and mathematical reasoning, 
    achieving $R^2 \approx 0.50$ on TruthfulQA and $R^2 \approx 0.40$ on 
    GSM8K across 20 independent model pairings.

    \item \textbf{Distributed Encoding:} Through ablation studies comparing 
    $L_2$ and $L_1$ regularization, we confirm that shared representations 
    are distributed rather than sparse. The collapse of $L_1$ mappers to 
    $R^2 \approx 0.01$ indicates semantic alignment relies holistically on 
    the full activation vector.

    \item \textbf{Architecture-Specific Intervention Dynamics:} We 
    demonstrate that optimal intervention strength is a property of the 
    student architecture rather than a universal hyperparameter, with student 
    models exhibiting characteristic sensitivity profiles that invert across 
    reasoning domains---Phi-3-Mini is the most steerable student on verbal 
    tasks and the least steerable on mathematical tasks.

    \item \textbf{Manifold Orthogonality:} We provide definitive evidence of 
    domain modularity via double dissociation confirmed across all 20 model 
    pairings without exception. Projection matrices trained on verbal 
    truthfulness collapse to a mean $R^2 = -3.83$ on arithmetic reasoning 
    and vice versa ($R^2 = -1.73$), providing strong evidence that distinct reasoning domains occupy 
    geometrically orthogonal latent subspaces.
\end{itemize}

\section{Related Work}
\label{sec:related}

\subsection{Linear Representations in Neural Networks}

The foundational observation motivating our work is that neural networks 
encode semantic knowledge geometrically. \citet{10.5555/2999792.2999959} 
first demonstrated this at scale in word embeddings, showing that semantic 
relationships manifest as stable linear directions in vector space. The 
Linear Representation Hypothesis, formalized by \citet{elhage2022solu} and 
validated by \citet{nanda2023emergentlinearrepresentationsworld}, extends 
this to transformer residual streams: semantic concepts are encoded as 
specific directions in activation space, making the model's internal state 
a structured geometric object whose distances and angles carry interpretable 
semantic content. If concepts are encoded as directions, a rotation and 
uniform scaling---operations captured exactly by a linear 
transformation---are in principle sufficient to align two coordinate systems 
encoding the same concepts in different orientations.

\subsection{Cross-Architecture Alignment and Model Stitching}

\citet{10.5555/3540261.3540279} demonstrated that 
independently trained models can form functional composites via model 
stitching: inserting a learned linear layer between the first $k$ layers of 
one network and the final $n-k$ layers of another produces surprisingly 
modest performance degradation, suggesting that a shared training objective 
is sufficient to produce compatible representations regardless of shared 
weights or structure. \citet{10.5555/3540261.3540693} extended this by 
developing rigorous metrics for representational similarity, showing that 
models converge to geometrically matchable representations even under 
different random initializations. Together, these results establish the 
\textit{affine compatibility assumption} as an empirically grounded prior. 
Our work departs from this foundation in a critical way: prior stitching 
routes activations through a fixed adapter at training time, whereas our 
method performs \textit{active inference-time intervention}, dynamically 
overwriting the student's residual stream during generation.

\subsection{Representation Engineering and Activation Steering}

A parallel body of work has demonstrated that model behavior can be steered 
by directly manipulating internal activations without weight updates. 
\citet{subramani2022extractinglatentsteeringvectors} showed that continuous 
steering vectors extracted from a model's residual stream can guide 
generation toward target behaviors. \citet{turner2024steeringlanguagemodelsactivation} 
demonstrated that fixed bias vectors added at specific layers reliably shift 
behavior across dimensions including honesty, sentiment, and persona. 
\citet{rimsky2023steeringllama2} extended this via contrastive activation 
addition, showing that difference-of-means vectors constructed from 
contrastive prompt pairs produce reliable behavioral shifts across tasks. 
\citet{stolfo2024improvinginstruction} further demonstrated that activation 
steering improves instruction-following without any weight update. 
\citet{zou2023universaltransferableadversarialattacks} revealed that such 
vectors exhibit universal properties, generalizing across model families and 
tasks. This literature establishes the residual stream as a writable 
medium, but all prior work operates within a single model. 
Our work extends this to cross-architecture steering, requiring 
explicit geometric translation before application.

Concurrently, \citet{wang2025expertsteer} propose ExpertSteer, which derives 
steering signals from an external expert model via autoencoder alignment and 
Recursive Feature Machines. Our method differs in two fundamental respects: 
we learn a single affine projection directly on paired hidden states rather 
than a nonlinear autoencoder pipeline, and we perform direct residual stream 
substitution rather than additive steering---a distinction that enables the 
geometric interpretability analysis central to our theoretical framework.

\subsection{Probing and Hierarchical Layer Processing}

Interpretability research consistently demonstrates that transformer layers 
process information hierarchically: lower layers resolve syntax, middle 
layers construct semantic representations, and final layers produce 
task-specific outputs \citep{48153, 
belinkov-2022-probing}. Because our teacher 
and student models differ substantially in depth, this hierarchy motivates 
parameterizing layer selection by \textit{relative processing depth} rather 
than absolute index---a layer at $75\%$ depth in a 32-layer model occupies 
an analogous functional role to one at $75\%$ depth in an 80-layer model. 
\citet{din2024jumpconclusionsshortcuttingtransformers} further establish 
injection timing as a consequential variable, directly motivating the 
temporal alignment strategy in Section~\ref{sec:temporal}.

\section{Theoretical Framework}
\label{sec:theoretical}

This section formalizes the three core components of our method: the affine 
compatibility assumption that justifies the projection, the intervention 
mechanics governing injection strength, and the temporal alignment strategy 
determining where in the student's pipeline injection must occur.

\subsection{Affine Manifold Compatibility}

Contemporary language models are trained to minimize next-token prediction 
loss over largely overlapping linguistic corpora. Because the statistical 
regularities of language are properties of the data rather than properties 
a model invents, two models compressing the same regularities will build 
geometrically compatible internal representations: their conceptual manifolds 
will align in topology even if coordinate orientations differ.

We formalize this as \textit{affine compatibility}. Rather than assuming 
strict isomorphism, we assume the relationship between the student's hidden 
state $h_S$ and the teacher's hidden state $h_T$ is well-approximated by:
\begin{equation}
    h_S \approx W h_T + b
    \label{eq:affine}
\end{equation}
where $W \in \mathbb{R}^{d_S \times d_T}$ captures the rotation and scaling 
required to align the teacher's conceptual manifold with the student's 
coordinate system, and $b$ absorbs mean shifts between spaces. We 
deliberately restrict this transformation to be linear: $W$ encodes a 
global rotation of the representational manifold, making every concept 
subject to the same geometric operation. A nonlinear projection would allow 
the adapter to warp different manifold regions independently, obscuring 
whether representations are genuinely compatible or merely forced into 
alignment by an expressive enough function.

\subsection{Intervention Strength and the $\alpha$ Spectrum}

We model the intervention as a combination of the student's original 
residual state and the projected teacher representation:
\begin{equation}
    h_{\text{final}} = (1 - \alpha)\,h_{\text{student}} + \alpha\,(W h_T)
    \label{eq:intervention}
\end{equation}
For $\alpha \in (0,1)$, this interpolates between the two states. At 
$\alpha = 1$, the student's state is replaced entirely by the projected 
teacher vector. For $\alpha > 1$, the formula enters an extrapolative 
regime whose geometric meaning is revealed by rearranging at $\alpha = 2$:
\begin{equation}
    h_{\text{final}} = W h_T + (W h_T - h_{\text{student}})
    \label{eq:extrapolation}
\end{equation}
Here the intervention simultaneously amplifies the projected teacher vector 
and subtracts the student's original activation---active error negation 
rather than blending. This is physically meaningful when the student's 
residual stream has accumulated sufficient inertia that simple substitution 
is insufficient to redirect computation 
\citep{nostalgebraist2020logit, elhage2021mathematical}. We treat $\alpha$ 
as an empirical quantity: different student architectures exhibit 
characteristic sensitivities to injected representations, and optimal 
$\alpha$ is a property of the student model rather than a universal 
hyperparameter. We note that Equation~\ref{eq:intervention} assumes the 
projected vector $Wh_T$ operates in a compatible magnitude regime; the 
practical realization of this constraint is detailed in 
Section~\ref{sec:methodology}.

\subsection{Temporal Alignment and Layer Selection}
\label{sec:temporal}

Effective intervention requires temporal alignment: the projected 
representation must arrive where it can propagate meaningfully to the output 
distribution, neither so early that subsequent layers overwrite it nor so 
late that decoding computation is bypassed. We parameterize layer selection 
by \textit{relative processing depth} $l \in [0, 1]$, mapping the 
hierarchical stages identified in the probing literature onto a common 
scale that transfers across architectures 
\citep{48153,
belinkov-2022-probing}.

We hypothesize a \textit{semantic handoff}: extracting from deep teacher 
layers ($l_T \approx 0.90$), where abstract reasoning is fully resolved, 
and injecting into mid-to-late student layers ($l_S \approx 0.75$), where 
the student transitions from semantic representation to vocabulary decoding 
\citep{din2024jumpconclusionsshortcuttingtransformers}. We treat relative 
depth as a principled \textit{search space} rather than a fixed prescription, 
as the optimal injection point is sensitive to each student architecture's 
functional organization. We therefore evaluate injection across a grid of 
relative depth combinations and characterize the empirical behavior in 
Section~\ref{sec:layer_analysis}.

\section{Methodology}
\label{sec:methodology}

\subsection{Datasets}

We evaluate on two benchmarks targeting distinct reasoning domains.
\textbf{TruthfulQA} \cite{inproceedings} is a 
dataset of 817 questions probing factual recall and verbal truthfulness 
across domains including health, law, history, and common misconceptions. 
Each question is paired with a \texttt{best\_answer} field and a 
\texttt{correct\_answers} field. We use the full validation split.
\textbf{GSM8K} \cite{cobbe2021gsm8k} is a dataset of grade school 
arithmetic word problems requiring multi-step numerical reasoning. We use 
the test split (1,319 questions) for intervention evaluation. Each problem 
is paired with a full chain-of-thought solution ending with a final numeric 
answer delimited by \texttt{\#\#\#\#}. We extract this final answer for 
evaluation. Both benchmarks use the prompt template 
\texttt{Question: \{question\} Answer:}, with \texttt{max\_new\_tokens=50} 
for TruthfulQA and \texttt{max\_new\_tokens=100} for GSM8K to accommodate 
numeric answer generation.

\subsection{Optimization Objective}
\label{sec:optimization}

For each model pair, we perform a full forward pass over all 
817 questions and extract the final-token hidden state at each 
candidate layer.
\begin{equation}
    \tilde{h} = \frac{h}{\|h\|_2}
    \label{eq:normalization}
\end{equation}
ensuring directional alignment is learned rather than magnitude differences, 
which are architectural artifacts. We optimize:
\begin{equation}
    \min_W \|H_S - H_T W\|_F^2 + \lambda\|W\|_F^2
    \label{eq:ridge}
\end{equation}
with fixed $\lambda = 0.1$, learned with a bias term corresponding to the 
full affine formulation in Equation~\ref{eq:affine}. All $R^2$ scores are 
reported on a held-out test set using a 70/30 train/test split (seed 42).

\subsection{Representation Structure Controls}

We employ two controls to validate that the learned $W$ captures genuine 
semantic geometry rather than spurious correlations.

\paragraph{Permutation Control.} A second Ridge regression is fit with 
rows of $H_S$ randomly shuffled, destroying semantic correspondence while 
preserving the marginal distribution. Any $R^2$ achieved by this sham 
mapper reflects distributional properties of the vectors rather than 
semantic alignment.

\paragraph{Sparsity Control ($L_1$).} We replace the $L_2$ penalty with 
an $L_1$ (Lasso) penalty ($\lambda = 0.0001$, max 5000 iterations):
\begin{equation}
    \min_W \|H_S - H_T W\|_F^2 + \lambda\|W\|_1
    \label{eq:lasso}
\end{equation}
If shared semantic information is localized to sparse features, the $L_1$ 
mapper should identify them and maintain competitive $R^2$. Collapse under 
sparsity would confirm that alignment relies on distributed correlations 
across the full activation vector.

\paragraph{Rotation Control (Orthogonal Procrustes).}
To test whether the full affine flexibility of Ridge regression is 
necessary or whether a rotation alone suffices, we fit an orthogonal 
Procrustes projection~\citep{schonemann1966generalized}: the rectangular 
generalization $W^* = UV^\top$ from the SVD of the cross-covariance 
matrix $H_S^\top H_T$, satisfying $W^{*\top} W^* = I_{d_S}$. 
This mapping is constrained to rotation only, with no scaling or shear. 
The rotation control collapses to mean $R^2 = -0.557$ on TruthfulQA and 
$R^2 = -1.290$ on GSM8K is substantially worse than even the permutation 
baseline ($R^2 = -0.274$ and $-0.218$ respectively). 
Cross-architecture latent spaces are therefore \textit{affinely} compatible 
but not \textit{rotationally} compatible: the scaling and bias components 
of the affine transformation capture systematic structural differences 
between architectures that a pure rotation cannot recover.

\subsection{Intervention Mechanics}
\label{sec:intervention_mechanics}

For each model pair and layer combination, we identify an 
\textit{opportunity set}: questions where the teacher answered correctly and 
the student did not. We pre-compute projected activations for all 
opportunity items:
\begin{equation}
    \hat{h}_S^{(i)} = W\tilde{h}_T^{(i)} + b
    \label{eq:projection}
\end{equation}
These are injected via a forward hook on the target layer, rescaled to 
match the $L_2$ norm of the student's current residual stream before 
application:
\begin{equation}
    h_{\text{final}} = (1-\alpha)\,h_{\text{student}} + 
    \alpha\,\bigl(\hat{h}_S \cdot \|h_{\text{student}}\|_2\bigr)
    \label{eq:intervention_full}
\end{equation}
ensuring the intervention operates as a directional correction regardless 
of dimensional gap between architectures. We sweep layer depths at four 
relative positions $l \in \{0.25, 0.50, 0.75, 0.90\}$ for both teacher 
and student (16 combinations per pairing) and the intervention coefficient 
across eight values:
\begin{equation}
    \alpha \in \{0.25,\ 0.5,\ 0.8,\ 1.0,\ 2.0,\ 3.0,\ 5.0,\ 10.0\}
    \label{eq:alpha_grid}
\end{equation}
All generation uses greedy decoding (\texttt{do\_sample=False}, max 50 
new tokens) ensuring fully deterministic outputs.

\subsection{Evaluation Metric}
\label{sec:evaluation}

We use deterministic matching rather than LLM-as-judge evaluation to avoid 
variance from probabilistic evaluators.

For \textbf{TruthfulQA}, success is binary inclusion against reference fields:
\begin{equation}
    \text{Score}(x) = \mathbb{I}(y_{\text{best}} \in x) \;\lor\; 
    \bigvee_{y \in \mathcal{Y}_{\text{correct}}} \mathbb{I}(y \in x)
    \label{eq:metric}
\end{equation}
where $x$ is the generated text lowercased and stripped.

For \textbf{GSM8K}, we extract the gold numeric answer from the 
\texttt{\#\#\#\#} delimiter and check for numeric equivalence in the generated 
text, handling formatting variants including comma separators and currency 
prefixes. The correction rate $\Delta$ reports the percentage of opportunity 
set items where the intervention converted an incorrect output to a correct 
one. This strict standard acts as a conservative lower bound: any observed 
$\Delta > 0$ is a direct consequence of the intervention forcing retrieval 
of the precise target concept.

\section{Experiments and Results}
\label{sec:experiments}

\subsection{Experimental Setup}
\label{sec:setup}

Table~\ref{tab:models} lists all models evaluated, spanning 
intra-family pairings, a domain-specialized teacher, and a 
synthetically trained student. The full sweep produces 2,560 
conditions: 20 pairings $\times$ 16 layer combinations 
$\times$ 8 $\alpha$ values.

\begin{table*}[t]
    \centering
    \small
    \begin{tabular}{llll}
        \toprule
        \textbf{Role} & \textbf{Model} & \textbf{Parameters} & 
        \textbf{Type} \\
        \midrule
        Teacher & \texttt{meta-llama/Meta-Llama-3-70B-Instruct} & 70B & 
        Dense \\
        Teacher & \texttt{meta-llama/Meta-Llama-3-8B-Instruct}  & 8B  & 
        Dense \\
        Teacher & \texttt{mistralai/Mistral-7B-Instruct-v0.3}   & 7B  & 
        MoE   \\
        Teacher & \texttt{google/gemma-2-9b-it}                 & 9B  & 
        Dense \\
        Teacher & \texttt{ibm-granite/granite-8b-code-instruct} & 8B  & 
        Code-specialized \\
        \midrule
        Student & \texttt{meta-llama/Llama-3.2-1B-Instruct}     & 1B  & 
        Dense \\
        Student & \texttt{Qwen/Qwen2.5-1.5B-Instruct}           & 1.5B & 
        Dense \\
        Student & \texttt{google/gemma-2-2b-it}                 & 2B  & 
        Dense \\
        Student & \texttt{microsoft/Phi-3-mini-4k-instruct}     & 3.8B & 
        Synthetic \\
        \bottomrule
    \end{tabular}
    \caption{Teacher and student models comprising the experimental matrix. 
    The Llama-70B teacher was loaded in 4-bit NF4 quantization due to 
    hardware constraints.}
    \label{tab:models}
\end{table*}

\subsection{Linear Alignment Efficacy}

When estimated via Ridge regression, the projection matrix consistently 
recovers substantial variance in the student's activation space. Across 
all 20 pairings and layer combinations, the Ridge mapper achieves mean 
$R^2 \approx 0.50$, with a peak of $R^2 = 0.684$ for the Llama-8B $\rightarrow$ 
Llama-1B intra-family pairing. We note that Llama-3.2-1B was 
trained with knowledge distillation from larger Llama models, 
which may independently inflate geometric compatibility for this 
specific pair beyond what the shared-distribution hypothesis 
alone would predict. The remaining 19 cross-family pairings 
share no such training relationship and still achieve 
$R^2 \geq 0.62$. Explaining approximately half 
the target variance via a single affine transformation across completely 
independent architectures with no shared weights, training procedure, or 
initialization constitutes a substantial structural signal.

The permutation control confirms that this signal reflects genuine semantic 
geometry. Destroying semantic correspondence by shuffling target vectors 
collapses regression to $R^2 \approx -0.22$ consistently across all 20 
pairings, ruling out exploitation of vector norms, distributional 
properties, or high-dimensional noise. The domain-specificity of this 
alignment is examined in Section~\ref{sec:dissociation}.

\subsection{Representation Structure: Distributed Encoding}
\label{sec:distributed}

Table~\ref{tab:regularization} reports mean $R^2$ under Ridge ($L_2$, 
$\lambda=0.1$), Lasso ($L_1$, $\lambda=0.0001$), and permutation control, 
averaged across all 20 pairings.

\begin{table*}[t]
    \centering
    \small
    \begin{tabular}{llcc}
        \toprule
        \textbf{Mapping Condition} & \textbf{Regularization} 
        & \textbf{TQA $R^2$} & \textbf{GSM $R^2$} \\
        \midrule
        Ridge Regression    & $L_2$ (distributed)          & $0.343$ & $0.256$ \\
        Procrustes Rotation & Orthogonal ($W^\top W = I$)  & $-0.557$ & $-1.290$ \\
        Lasso Regression    & $L_1$ (sparse)               & $0.014$ & --- \\
        Permutation Control & $L_2$ (shuffled)             & $-0.274$ & $-0.218$ \\
        \bottomrule
    \end{tabular}
    \caption{Projection performance under different mapping conditions,
    averaged across all 20 model pairings at $l_T = l_S = 0.75$.
    The orthogonal Procrustes ablation shows that
    rotation-only mapping collapses below the permutation baseline,
    confirming that the scaling and bias components of the affine
    fit are load-bearing.}
    \label{tab:regularization}
\end{table*}

The $L_1$ mapper flatlines at $R^2 \approx 0.014$---barely above the 
permutation baseline and a near-total collapse from the Ridge mapper. 
Forcing sparsity destroys the mapping entirely, confirming that 
cross-architecture alignment relies on the full distributed structure 
of the activation vector and cannot be recovered from any sparse subset 
of dimensions. This directly validates the use of dense Ridge regression 
as the appropriate projection estimator.

\subsection{Downstream Intervention Efficacy}
\label{sec:intervention_efficacy}

\begin{figure}[h]
    \centering
    \includegraphics[width=\linewidth]{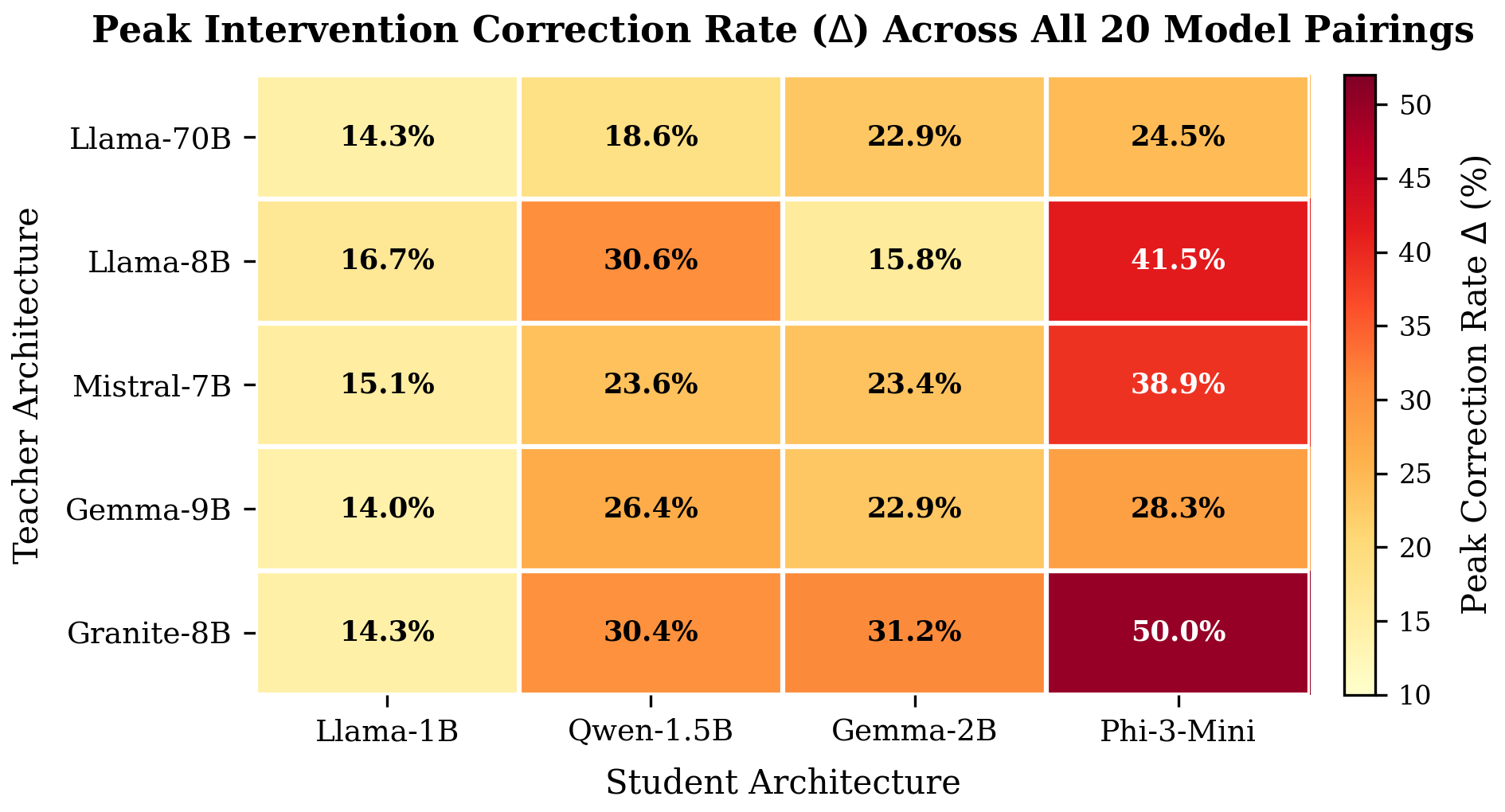}
    \caption{Peak intervention correction rate ($\Delta$) across the full 
    $5 \times 4$ teacher-student matrix. Each cell reports the maximum 
    $\Delta$ over all layer combinations and $\alpha$ values.}
    \label{fig:heatmap}
\end{figure}

Figure~\ref{fig:heatmap} presents peak correction rate $\Delta$ across all 
20 model pairings. The intervention is effective without exception, 
demonstrating that cross-architecture latent steering generalizes across 
heterogeneous model families, parameter scales, and training methodologies. 
Correction rates range from $14.0\%$ (Gemma-9B $\rightarrow$ Llama-1B) 
to $50.0\%$ (Granite-8B $\rightarrow$ Phi-3-Mini), with a mean of $25.2\%$.

Phi-3-Mini is consistently the most steerable student, receiving the 
highest correction rate from four of five teachers. We attribute this to 
its synthetic training corpus: curated, structured training data may 
produce more geometrically regular internal representations amenable to 
external redirection. The Llama-70B teacher consistently underperforms 
Llama-8B despite its larger parameter count, a result of two compounding 
factors: the dimensional reduction from $d_T = 8192$ to $d_S \leq 2048$ 
introduces lossy compression, and 4-bit quantization injects rounding 
noise into extracted activation vectors, degrading projection precision.

\paragraph{Geometric Alignment Does Not Predict Behavioral 
Correction.} The near-zero correlation ($r = -0.071$) between 
$R^2$ and $\Delta$ reveals a fundamental dissociation: 
directional accuracy matters more than variance explained. 
A projection pointing in the right direction outperforms 
a high-$R^2$ projection that fits the full distributional 
structure of the target space.

\subsection{Architectural Sensitivity: Alpha and Layer Analysis}
\label{sec:alpha_analysis}\label{sec:layer_analysis}

\begin{figure}[h]
    \centering
    \includegraphics[width=\linewidth]{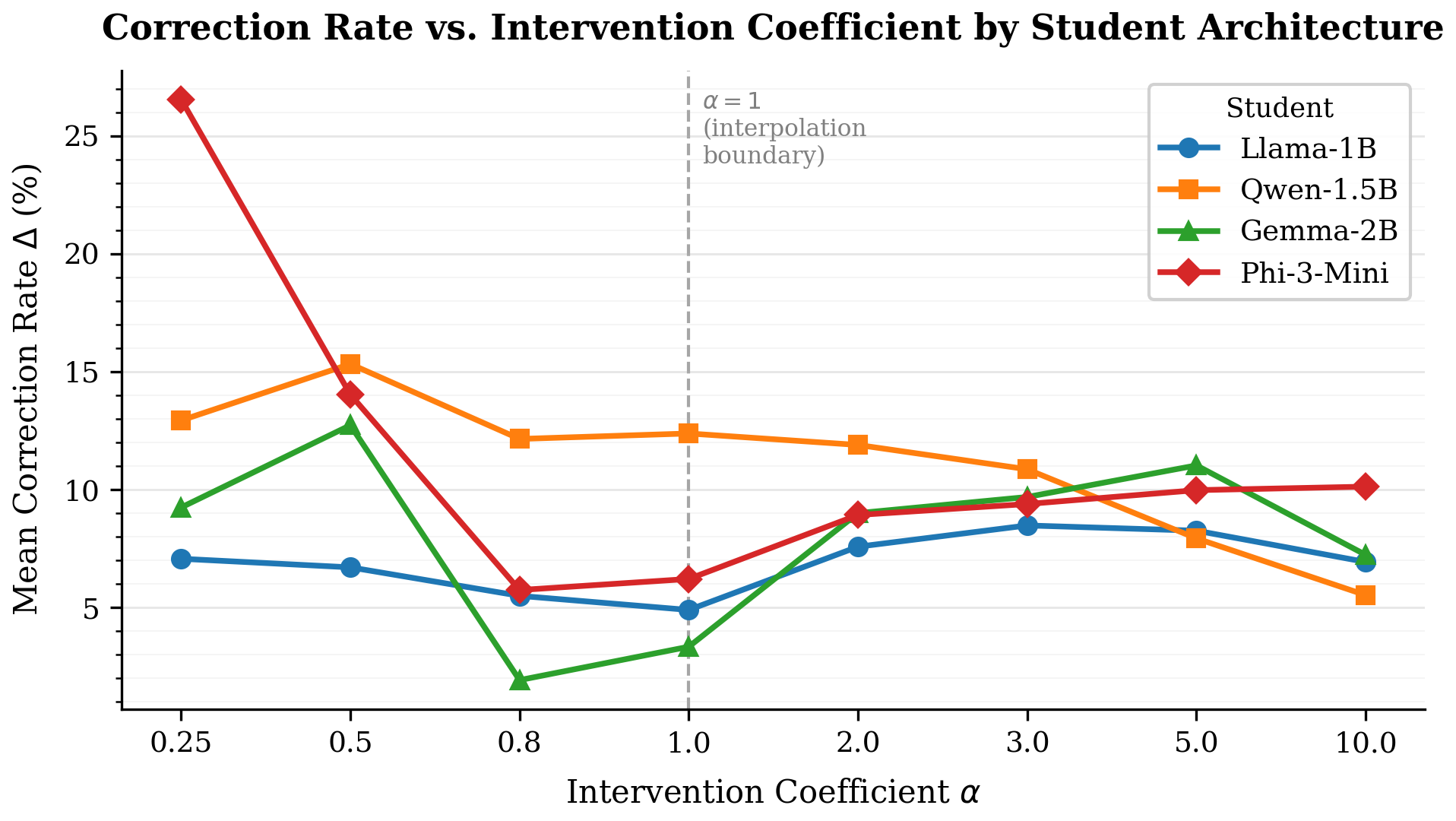}
    \caption{Mean $\Delta$ as a function of $\alpha$ per student 
    architecture, averaged across all teachers and layer combinations. 
    The dashed line marks $\alpha = 1$. Phi-3-Mini exhibits a sharp peak 
    at $\alpha = 0.25$ with immediate collapse; Llama-1B maintains a flat 
    response across the sweep.}
    \label{fig:alpha}
\end{figure}

\paragraph{Intervention Coefficient.}
Figure~\ref{fig:alpha} plots mean correction rate vs.\ $\alpha$ per 
student architecture. Rather than a universal optimum, the four 
architectures exhibit markedly distinct sensitivity profiles.

Phi-3-Mini peaks sharply at $\alpha = 0.25$ (mean $\Delta = 26.5\%$) 
before collapsing below $10\%$ for all $\alpha \geq 0.8$, indicating high 
sensitivity to representational displacement---a gentle directional nudge 
suffices while stronger interventions destabilize generation. Llama-1B 
exhibits the flattest profile, tolerating a broad injection range, 
suggesting greater representational inertia. Across all architectures, 
optimal performance occurs within the interpolation regime ($\alpha \leq 1$) 
in $80.6\%$ of configurations (median optimal $\alpha = 0.5$). 
Extrapolative coefficients account for the remaining $19.4\%$, 
concentrated in Llama-1B. A consistent dip at $\alpha \in \{0.8, 1.0\}$ 
across multiple architectures suggests full replacement of the student's 
residual state is frequently counterproductive; partial blending 
outperforms complete substitution for most models.

\begin{figure}[h]
    \centering
    \includegraphics[width=0.78\linewidth]{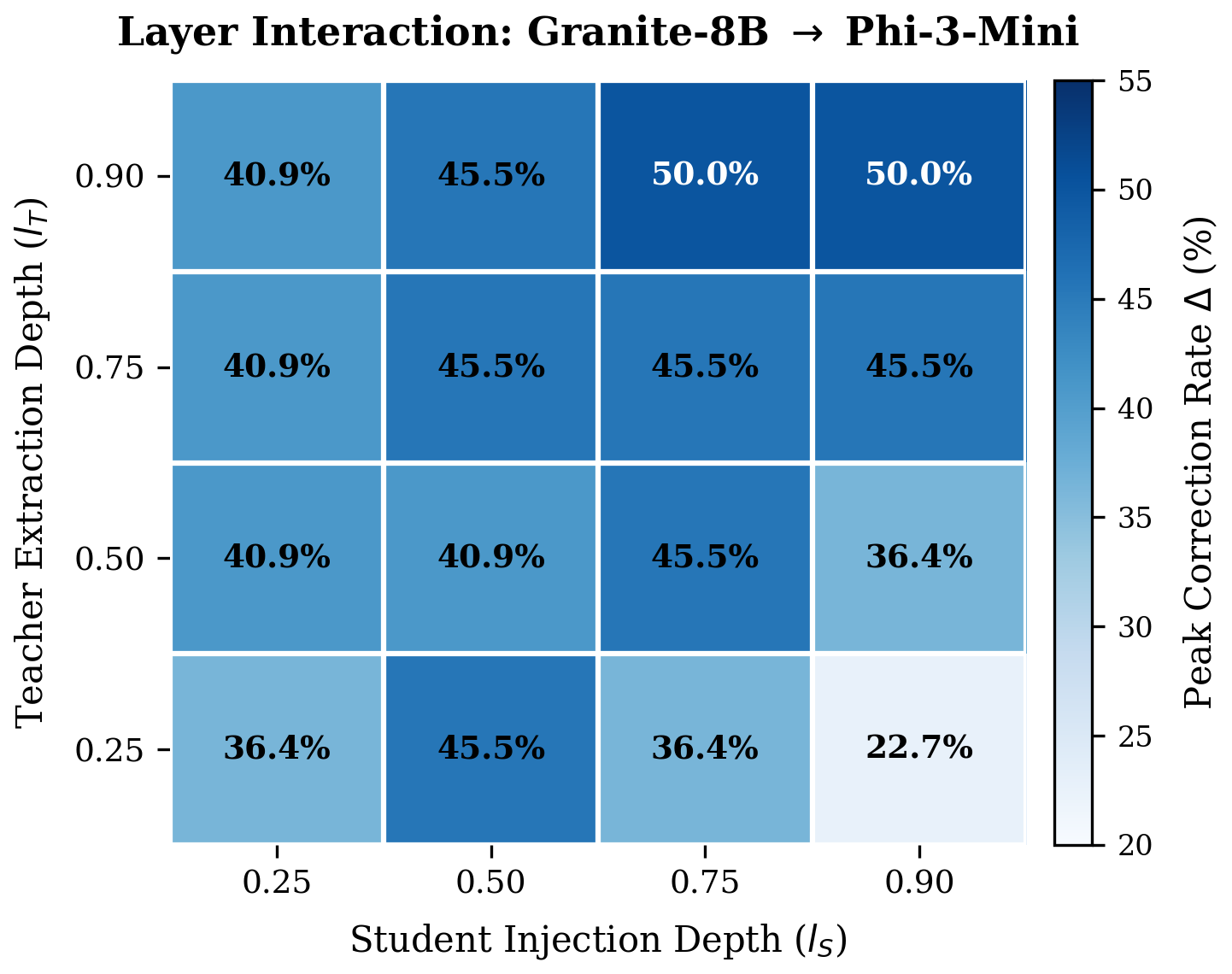}
    \caption{Peak $\Delta$ across 16 $l_T \times l_S$ combinations for 
    Granite-8B $\rightarrow$ Phi-3-Mini. Performance increases 
    monotonically with teacher extraction depth; student injection 
    depth shows a weaker, less consistent effect.}
    \label{fig:layers}
\end{figure}

\paragraph{Layer Selection.}
Figure~\ref{fig:layers} presents the layer interaction heatmap for 
Granite-8B $\rightarrow$ Phi-3-Mini.

Correction rates increase monotonically as $l_T$ increases from $0.25$ 
to $0.90$, consistent with the semantic handoff hypothesis: deeper teacher 
layers yield more fully resolved conceptual representations. Student 
injection depth shows a weaker pattern; peak cells at $l_T = 0.90$ are 
achieved at both $l_S = 0.75$ and $l_S = 0.90$. This asymmetry 
establishes deep teacher extraction as a more reliable design principle 
than any specific student injection target.

\subsection{Generalization to Mathematical Reasoning}
\label{sec:gsm8k}

\begin{figure}[h]
    \centering
    \includegraphics[width=\linewidth]{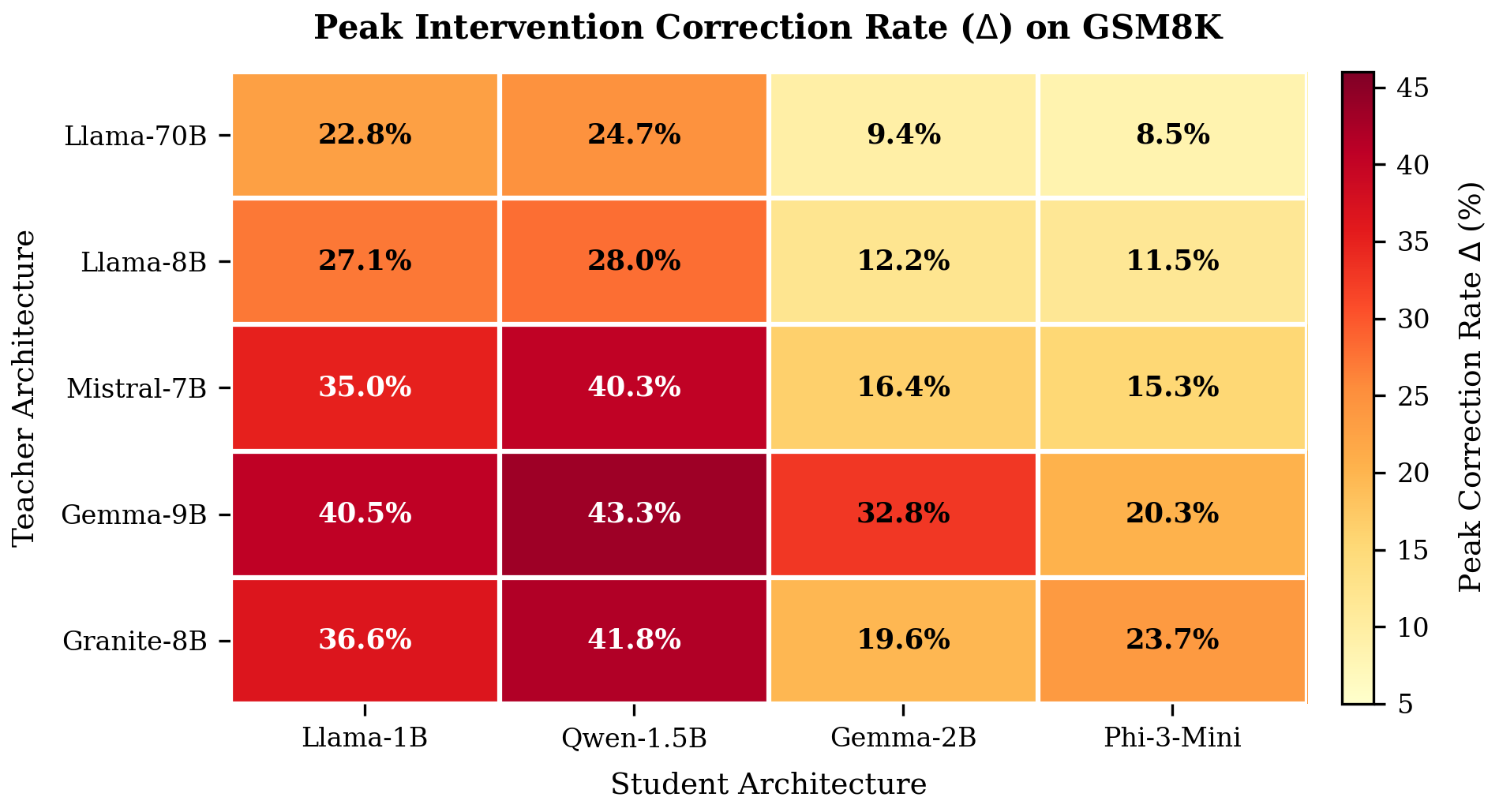}
    \caption{Peak intervention correction rate ($\Delta$) on GSM8K across 
    the full $5 \times 4$ teacher-student matrix. Correction rates range 
    from $8.5\%$ to $43.3\%$ with a mean of $25.5\%$, virtually identical 
    to the TruthfulQA mean of $25.2\%$.}
    \label{fig:heatmap_gsm}
\end{figure}

To assess whether cross-architecture steering generalizes beyond verbal 
reasoning, we replicate the full intervention sweep on GSM8K, evaluating 
all 20 model pairings across the identical grid of layer combinations and 
$\alpha$ values. Figure~\ref{fig:heatmap_gsm} presents peak correction 
rates across the model matrix.

The method generalizes completely: correction rates range from $8.5\%$ to 
$43.3\%$ with a mean of $25.5\%$, statistically indistinguishable from the 
TruthfulQA mean of $25.2\%$. This replication across a fundamentally 
different cognitive domain --- multi-step arithmetic versus short-form 
factual recall --- establishes cross-architecture latent steering as a 
general-purpose correction mechanism rather than a benchmark-specific 
artifact.

Two structural patterns distinguish the GSM8K results from TruthfulQA. 
First, student steerability rankings invert completely across domains. 
Phi-3-Mini, the most steerable student on TruthfulQA (peak $\Delta = 50.0\%$), 
becomes the least steerable on GSM8K (peak $\Delta = 23.7\%$). Conversely, 
Llama-1B, the hardest student to steer verbally (peak $\Delta = 16.7\%$), 
becomes one of the most steerable mathematically (peak $\Delta = 40.5\%$). 
This complete rank reversal demonstrates that steerability is not an 
intrinsic architectural property but a domain-specific one, reflecting the 
geometric organization of each model's reasoning subspaces rather than its 
overall representational plasticity.

Second, the optimal $\alpha$ collapses entirely to the sub-interpolation 
regime on GSM8K: $\alpha = 0.25$ is optimal in $99.4\%$ of configurations, 
and mean correction rate hits $0.0\%$ for all $\alpha \geq 0.8$. 
Mathematical representations are substantially more fragile than verbal ones 
--- a gentle directional nudge suffices, while any stronger intervention 
destroys generation entirely. This asymmetry suggests that mathematical 
reasoning subspaces are more narrowly organized, with less tolerance for 
representational displacement before coherent output collapses.

\subsection{Domain Modularity: A Universal Double Dissociation}
\label{sec:dissociation}

We now ask whether learned geometry transfers across domains. 
For each of the 20 model pairings at $l_T = l_S = 0.75$, we 
train a projection on 200 in-domain questions and evaluate it 
on 100 held-out questions from the opposite domain, in both 
directions.

\begin{table}[t]
    \centering
    \small
    \setlength{\tabcolsep}{4pt}
    \begin{tabular}{lcccc}
        \toprule
        & \textbf{TQA} & \textbf{TQA$\to$GSM} 
        & \textbf{GSM} & \textbf{GSM$\to$TQA} \\
        \midrule
        Mean $R^2$ & $0.343$ & $-3.831$ & $0.256$ & $-1.729$ \\
        Min $R^2$  & $0.278$ & $-8.170$ & $0.176$ & $-3.227$ \\
        Max $R^2$  & $0.446$ & $-1.807$ & $0.351$ & $-1.081$ \\
        \bottomrule
    \end{tabular}
    \caption{Double dissociation across all 20 pairings. 
    In-domain columns show held-out $R^2$; transfer columns 
    show cross-domain $R^2$. Both directions collapse 
    catastrophically. Confirmed 20/20.}
    \label{tab:dissociation}
\end{table}

The results constitute a universal double dissociation confirmed without 
exception across all 20 pairings. The TruthfulQA-trained mapper collapses 
to a mean $R^2 = -3.831$ on GSM8K; the GSM8K-trained mapper collapses to 
mean $R^2 = -1.729$ on TruthfulQA. Both transferred projections perform 
catastrophically worse than predicting the target mean ($R^2 = 0$), 
indicating active geometric interference rather than mere uninformativeness. 
The worst single cell reaches $R^2 = -8.170$ (Granite-8B $\rightarrow$ 
Qwen-1.5B, TQA-trained mapper applied to GSM8K), meaning the projection 
orients student activations in a direction nearly maximally opposed to 
correct mathematical representations.

Notably, Qwen-1.5B consistently shows the most extreme
dissociation in both directions (TQA$\to$GSM range:
$-7.867$ to $-8.170$; GSM$\to$TQA range: $-2.908$
to $-3.227$), approximately twice the interference
magnitude of other students, suggesting its reasoning
subspaces are more modularly organized with sharper
geometric boundaries between domains.

To verify that this finding is not specific to the fixed 
layer combination used in the 20-pair sweep, we conduct a complementary experiment 
sweeping all 16 layer depth combinations for the Mistral-7B $\rightarrow$ 
Qwen-1.5B pair, evaluating dissociation at every $l_T \times l_S$ 
intersection. The dissociation is confirmed in all 16/16 combinations 
(Appendix~\ref{app:layer_dissociation}), ruling out the possibility that 
orthogonality is a depth-specific artifact. The magnitude of interference 
varies systematically with injection depth: mid-network injection 
($l_S \approx 0.50$) produces the most catastrophic transfer collapse 
(TQA$\rightarrow$GSM mean $R^2 = -11.5$), while late injection 
($l_S \approx 0.90$) produces the weakest interference 
(TQA$\rightarrow$GSM mean $R^2 = -3.1$). This gradient confirms that 
mid-network layers are the primary site of domain-specific semantic 
construction --- mismatched representations injected there corrupt the 
entire downstream computation, while late injection gives the network 
less time to propagate the interference before output.

These results establish domain orthogonality as a consistent 
geometric property confirmed across all 20 tested pairings: 
a projection matrix must be trained on data representative 
of the target reasoning domain.

\section{Conclusion}

We have presented a framework for cross-architecture inference-time 
steering via learned linear projection of latent representations. A 
single affine transformation suffices to translate activation vectors 
between heterogeneous architectures, and injecting translated 
representations into a student model's residual stream produces reliable 
behavioral corrections without weight updates, shared architecture, or 
access to training data.

The method generalizes across reasoning domains, achieving mean 
correction rates of $25.2\%$ on TruthfulQA and $25.5\%$ on GSM8K 
across 20 heterogeneous pairings. Four findings sharpen the picture: 
geometric fidelity does not predict behavioral correction 
($r \approx -0.07$); optimal intervention strength inverts across 
domains; deep teacher extraction ($l_T \approx 0.90$) is a more 
reliable design principle than any specific student injection target; 
and mathematical representations tolerate substantially less 
representational displacement than verbal ones.

Finally, the double dissociation between verbal and mathematical 
domains, confirmed across all 20 model pairings without exception 
and across all 16 layer combinations for the pilot pair, establishes domain-specific subspace geometry as a consistent 
property confirmed across all 20 tested pairings rather than 
an artifact of any particular architecture. A projection matrix 
must be trained on data representative of the target reasoning domain, 
and the geometric structure of one reasoning type cannot be repurposed 
for another.

\section*{Limitations}

\paragraph{Benchmark Scope.} All behavioral experiments use 
instruction-tuned models with short-form generation (50--100 
tokens), evaluated on TruthfulQA and GSM8K. This scope 
excludes two important directions. First, dedicated reasoning 
models like DeepSeek-R1, Phi-4-reasoning, and Qwen3-Thinking generate 
extended chain-of-thought traces before producing a final answer, 
and it is an open question whether residual-stream substitution 
at a single layer remains effective when a long generated trace 
occupies the context. The intervention may need to be applied 
earlier, at multiple layers, or in coordination with the 
reasoning trace rather than independently of it. Second, 
correction rates on commonsense reasoning, causal inference, 
multi-step planning, and longer-form generation have not been 
established, and the domain dissociation result does not 
speak to whether additional reasoning types would occupy 
distinct or overlapping latent subspaces.

\paragraph{Sparse Layer Grid.}
Layer selection is evaluated at only four relative depth positions, leaving 
behavior between grid points uncharacterized. Finer-grained sweeps may 
reveal non-monotonic dynamics or narrow optimal windows the current 
resolution cannot detect.

\paragraph{Quantization Noise.}
The Llama-70B teacher was loaded in 4-bit NF4 quantization due to hardware 
constraints, introducing rounding noise into extracted activations. The 70B 
results represent a lower bound on full-precision performance rather than 
a representative evaluation of large-scale teacher utility.

\paragraph{Dissociation Layer Generalization.}
The full layer sweep confirming dissociation across all 16 depth 
combinations is conducted on a single model pair (Mistral-7B 
$\rightarrow$ Qwen-1.5B). Whether the systematic variation in interference 
magnitude with injection depth generalizes across other architecture 
combinations remains unverified, though the binary dissociation result 
itself is confirmed across all 20 pairs at a representative layer combination.

\paragraph{Open-Weight Models Only.}
Whether affine compatibility holds for proprietary closed-weight 
architectures or models trained on substantially different data 
distributions remains untested.

\paragraph{Evaluation Metric Ceiling.}
Deterministic substring matching counts interventions that redirect 
reasoning but produce lexically distinct correct answers as failures. 
Reported $\Delta$ values are conservative lower bounds on true 
behavioral impact.

\section*{Ethical Considerations}

This work investigates the geometric structure of latent representations 
in large language models and demonstrates a method for correcting model 
outputs at inference time via cross-architecture activation steering.

We identify two areas warranting ethical consideration. First, the 
intervention mechanism demonstrated here---which modifies a model's 
internal reasoning trajectory without any weight update or user-visible 
indication---could in principle be applied to steer model outputs in 
undesirable directions as readily as beneficial ones. We emphasize that 
all experiments in this work are directed at \textit{correcting} factually 
incorrect outputs toward ground-truth answers, and that the method requires 
explicit access to the target model's internal activations, limiting its 
applicability outside controlled research settings. Second, all models used 
in this work are publicly available open-weight models. No proprietary 
systems, private data, or human subjects were involved in any experiment. 
TruthfulQA and GSM8K are publicly available benchmarks with no sensitive 
personal information.

We believe the primary contribution of this work---characterizing the 
geometric structure of shared latent representations and their domain 
specificity---is a positive contribution to the interpretability and 
transparency of large language models.

\section*{Reproducibility Statement}

We have made substantial efforts to ensure the reproducibility of all 
results reported in this paper. All experiments use publicly available 
open-weight models accessible via HuggingFace, and all benchmarks 
(TruthfulQA, GSM8K) are publicly available datasets. Exact model 
identifiers, quantization configurations, random seeds, train/test split 
ratios, regularization hyperparameters, and generation settings are 
specified in Section~\ref{sec:methodology} and 
Appendix~\ref{app:implementation}. The intervention hook implementation, 
projection matrix estimation procedure, and evaluation logic are described 
in full in Appendix~\ref{app:implementation}. The layer dissociation 
experiment parameters---sample sizes, layer combinations, and Ridge 
regression configuration---are specified in 
Section~\ref{sec:dissociation} and Appendix~\ref{app:layer_dissociation}. 
All results are deterministic: generation uses greedy decoding 
(\texttt{do\_sample=False}) throughout, and all random operations use 
\texttt{random\_state=42}. Complete experimental code will be released as 
anonymous supplementary material with the submission.


\bibliography{references} 


\appendix

\section{Full Geometric Alignment Results}
\label{app:r2}

\begin{figure}[h]
    \centering
    \includegraphics[width=\linewidth]{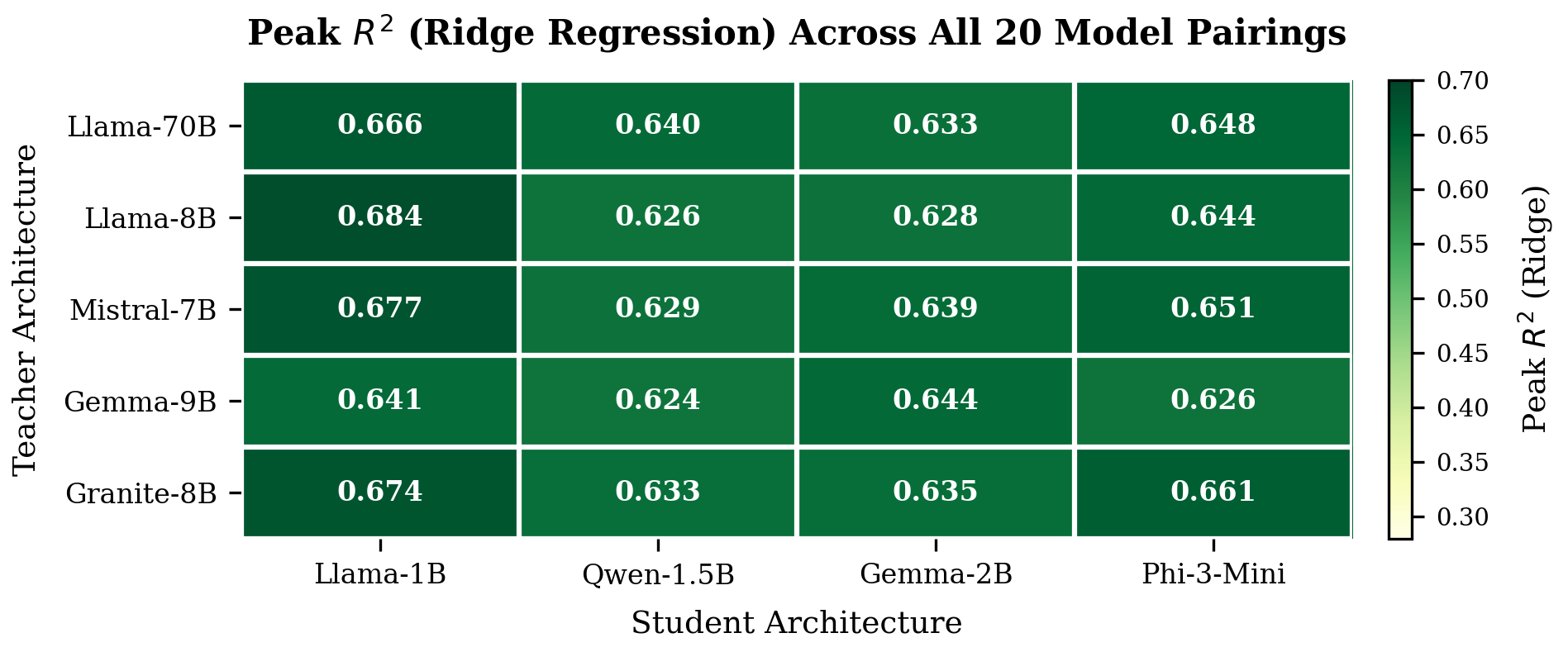}
    \caption{Peak $R^2$ (Ridge regression) for each of the 20 
    teacher-student pairings, maximized over all 16 layer combinations. 
    The Llama-8B $\rightarrow$ Llama-1B intra-family pairing achieves 
    the highest alignment ($R^2 = 0.684$). All pairings exceed 
    $R^2 = 0.62$, confirming broad affine compatibility across 
    the experimental matrix.}
    \label{tab:r2_full}
\end{figure}

Table~\ref{tab:r2_full} reports the peak $R^2$ achieved by the Ridge 
projection for each of the 20 teacher-student pairings, maximized over 
all 16 layer depth combinations. Values are consistently in the range 
$[0.62, 0.68]$ across the five teachers, with the notable exception of 
Gemma-9B pairings which cluster slightly lower at $[0.62, 0.64]$. The 
absence of any pairing with $R^2 < 0.60$ confirms that affine 
compatibility holds broadly across the experimental matrix and is not 
dependent on architectural similarity between teacher and student.

\section{Extended Layer Interaction Heatmaps}
\label{app:layers}

\begin{figure}[h]
    \centering
    \includegraphics[width=\linewidth]{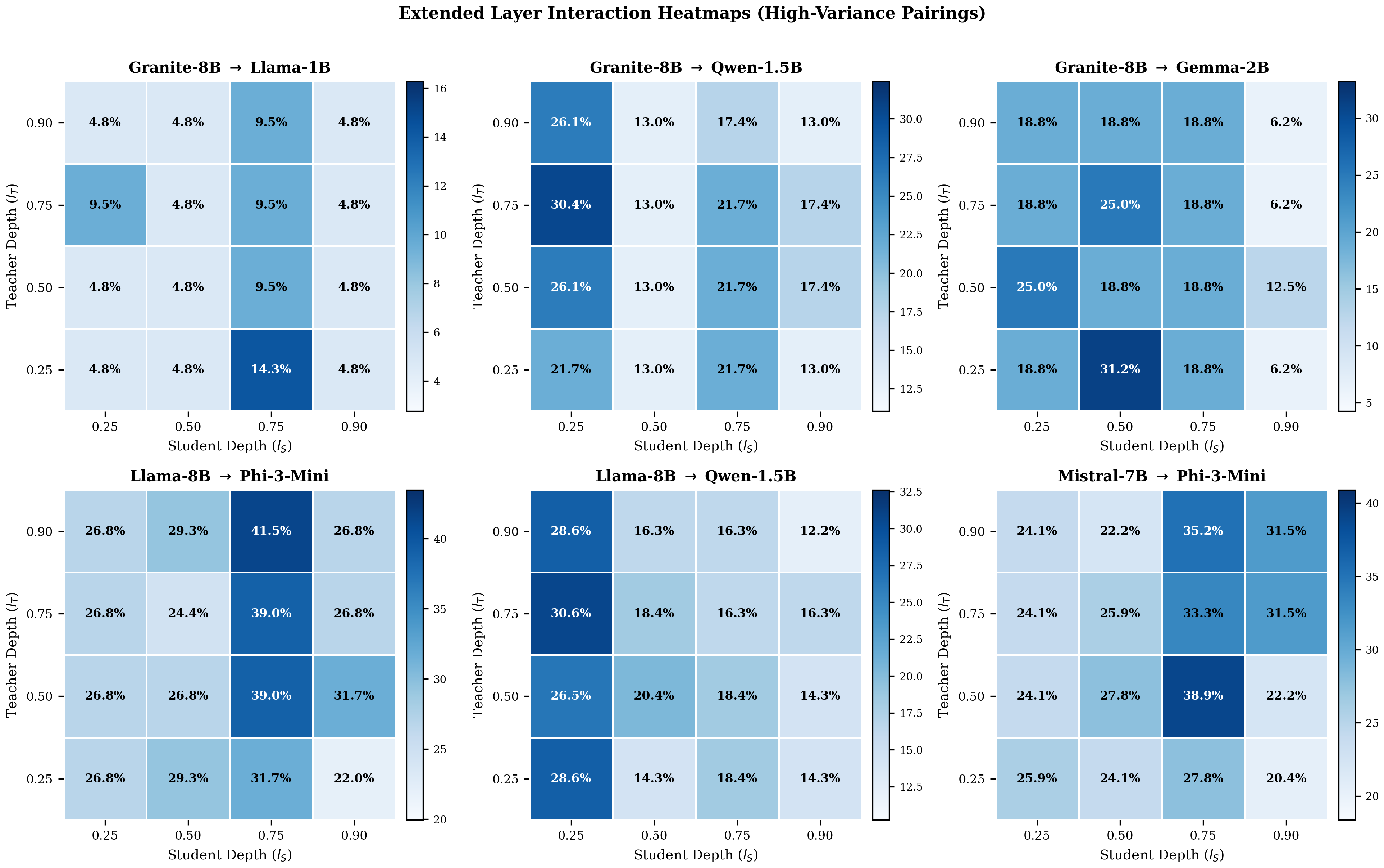}
    \caption{Peak correction rate $\Delta$ across all 16 $l_T \times 
    l_S$ combinations for the six highest-variance model pairings. 
    Rows are ordered with deepest teacher extraction at the top. The 
    monotonic improvement with teacher extraction depth observed in the 
    main paper (Figure~\ref{fig:layers}) generalizes across these 
    pairings, while student injection depth consistently shows a weaker 
    and less structured effect. The Granite-8B $\rightarrow$ Gemma-2B 
    pairing shows the highest within-pair variance ($\sigma = 6.7\%$), 
    with a pronounced performance gradient that makes layer selection 
    particularly consequential for this pairing.}
    \label{fig:app_layers}
\end{figure}

Figure~\ref{fig:app_layers} presents layer interaction heatmaps for the 
six model pairings with the highest within-pair variance in correction 
rate across layer combinations (standard deviation $> 5\%$). These 
pairings were selected because their heatmaps are most informative: 
the layer choice meaningfully affects outcome, and the gradient patterns 
visible in each provide evidence about the generalizability of the 
semantic handoff hypothesis discussed in Section~\ref{sec:temporal}.

\section{Per-Teacher Alpha Sensitivity Profiles}
\label{app:alpha}

\begin{figure}[h]
    \centering
    \includegraphics[width=\linewidth]{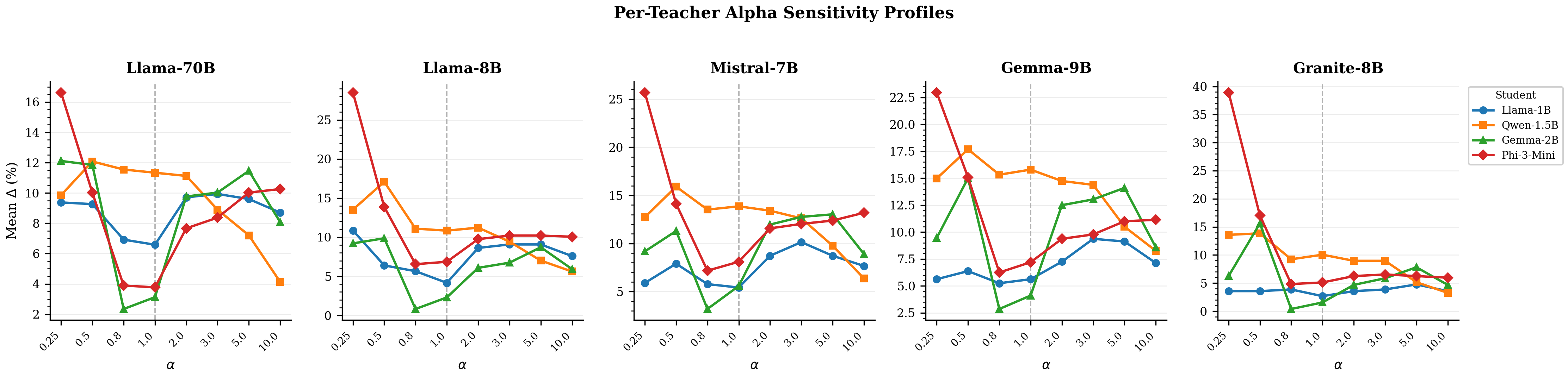}
    \caption{Mean correction rate $\Delta$ vs.\ $\alpha$ broken down by 
    teacher architecture, with one line per student. The dashed vertical 
    line marks $\alpha = 1$. Student sensitivity profiles are highly 
    consistent across teachers: Phi-3-Mini peaks at $\alpha = 0.25$ and 
    collapses immediately regardless of teacher, while Qwen-1.5B 
    consistently prefers $\alpha = 0.5$. The Mistral-7B panel is the 
    notable exception, where Gemma-2B achieves its best performance at 
    $\alpha = 2.0$ rather than the sub-interpolation values preferred 
    across other teachers, suggesting a specific compatibility between 
    Mistral's MoE representations and Gemma-2B's residual stream that 
    benefits from stronger intervention.}
    \label{fig:app_alpha}
\end{figure}

Figure~\ref{fig:app_alpha} decomposes the alpha sensitivity curves from 
Figure~\ref{fig:alpha} by teacher architecture, plotting mean correction 
rate vs.\ $\alpha$ for each student separately within each teacher panel. 
This view addresses a question the main paper leaves open: whether the 
characteristic sensitivity profiles of each student are consistent across 
teachers, or whether the teacher architecture modulates them.

\section{Full Double Dissociation Results}
\label{app:dissociation}

Table~\ref{tab:app_dissociation} reports all four $R^2$ values for each 
of the 20 teacher-student pairings in the domain dissociation experiment. 
In-domain values reflect proper held-out evaluation on 100 test questions. 
Transfer values confirm catastrophic collapse in both directions for every 
pairing without exception.

\begin{table*}[t]
    \centering
    \small
    \begin{tabular}{llcccc}
        \toprule
        \textbf{Teacher} & \textbf{Student} & \textbf{TQA} & 
        \textbf{TQA$\rightarrow$GSM} & \textbf{GSM} & 
        \textbf{GSM$\rightarrow$TQA} \\
        \midrule
        Llama-70B  & Llama-1B   & $0.354$ & $-2.748$ & $0.198$ & $-1.108$ \\
        Llama-70B  & Qwen-1.5B  & $0.368$ & $-8.133$ & $0.222$ & $-3.005$ \\
        Llama-70B  & Gemma-2B   & $0.290$ & $-2.838$ & $0.296$ & $-1.261$ \\
        Llama-70B  & Phi-3-Mini & $0.285$ & $-1.807$ & $0.300$ & $-1.567$ \\
        Llama-8B   & Llama-1B   & $0.389$ & $-2.389$ & $0.230$ & $-1.101$ \\
        Llama-8B   & Qwen-1.5B  & $0.372$ & $-7.867$ & $0.226$ & $-3.037$ \\
        Llama-8B   & Gemma-2B   & $0.290$ & $-2.798$ & $0.303$ & $-1.182$ \\
        Llama-8B   & Phi-3-Mini & $0.296$ & $-1.834$ & $0.254$ & $-1.529$ \\
        Mistral-7B & Llama-1B   & $0.377$ & $-2.439$ & $0.244$ & $-1.081$ \\
        Mistral-7B & Qwen-1.5B  & $0.405$ & $-8.078$ & $0.176$ & $-3.227$ \\
        Mistral-7B & Gemma-2B   & $0.312$ & $-2.714$ & $0.297$ & $-1.178$ \\
        Mistral-7B & Phi-3-Mini & $0.291$ & $-1.827$ & $0.220$ & $-1.670$ \\
        Gemma-9B   & Llama-1B   & $0.353$ & $-2.748$ & $0.190$ & $-1.176$ \\
        Gemma-9B   & Qwen-1.5B  & $0.398$ & $-8.019$ & $0.226$ & $-2.908$ \\
        Gemma-9B   & Gemma-2B   & $0.361$ & $-2.513$ & $0.326$ & $-1.115$ \\
        Gemma-9B   & Phi-3-Mini & $0.303$ & $-1.870$ & $0.301$ & $-1.527$ \\
        Granite-8B & Llama-1B   & $0.417$ & $-2.845$ & $0.195$ & $-1.191$ \\
        Granite-8B & Qwen-1.5B  & $0.446$ & $-8.170$ & $0.275$ & $-2.922$ \\
        Granite-8B & Gemma-2B   & $0.281$ & $-3.060$ & $0.289$ & $-1.313$ \\
        Granite-8B & Phi-3-Mini & $0.278$ & $-1.930$ & $0.351$ & $-1.489$ \\
        \midrule
        \textit{Mean} & & $0.343$ & $-3.831$ & $0.256$ & $-1.729$ \\
        \bottomrule
    \end{tabular}
    \caption{Full double dissociation results for all 20 pairings. 
    TQA and GSM columns show held-out in-domain $R^2$. 
    TQA$\rightarrow$GSM and GSM$\rightarrow$TQA show transfer $R^2$. 
    Dissociation confirmed in all 20 pairs.}
    \label{tab:app_dissociation}
\end{table*}

\section{GSM8K Alpha Sensitivity Profiles}
\label{app:gsm_alpha}

\begin{figure}[h]
    \centering
    \includegraphics[width=\linewidth]{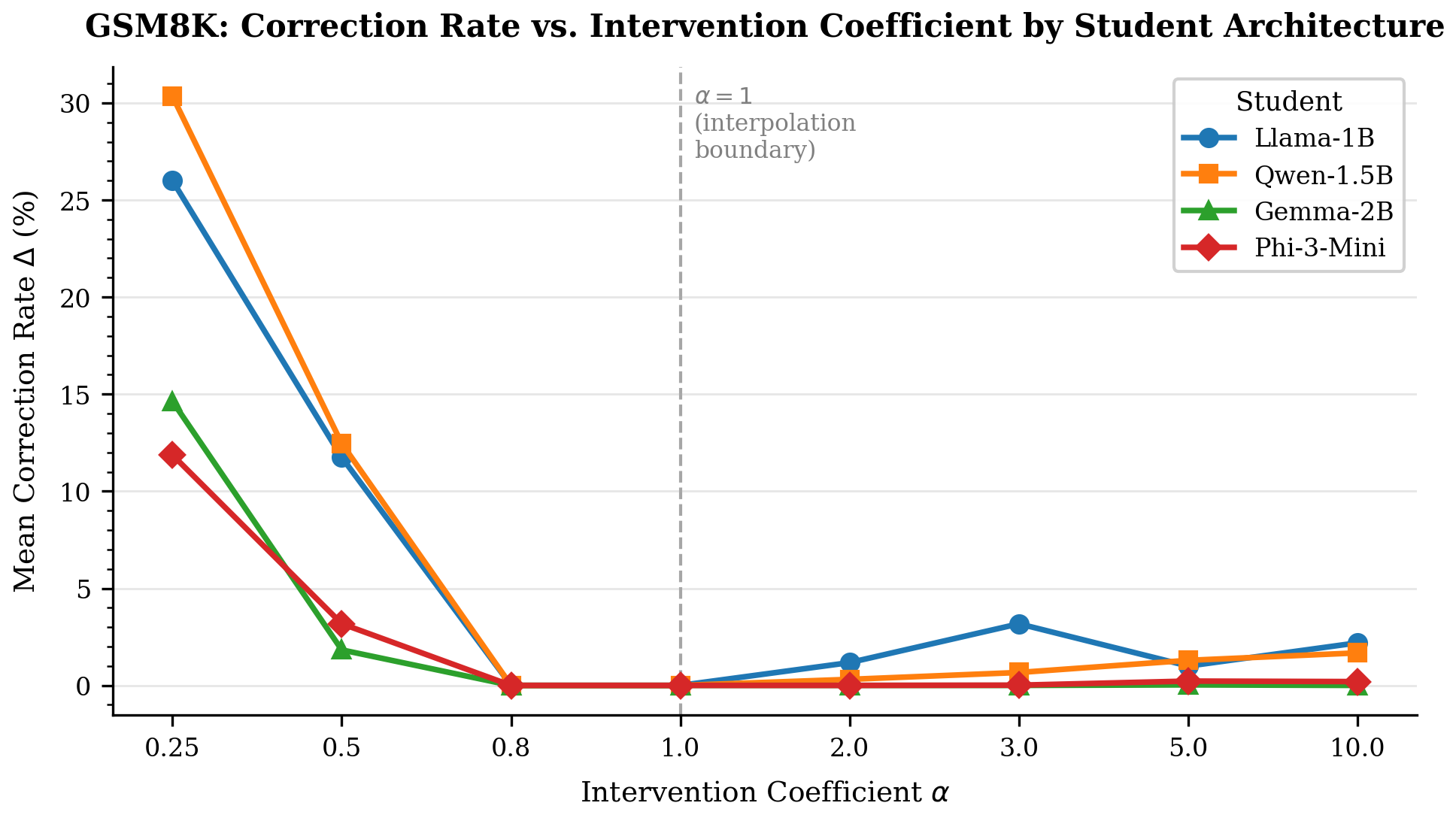}
    \caption{Mean correction rate $\Delta$ vs.\ $\alpha$ on GSM8K per 
    student architecture, averaged across all teachers and layer 
    combinations. All architectures collapse to near-zero for 
    $\alpha \geq 0.8$, in sharp contrast to TruthfulQA profiles where 
    partial recovery occurs at higher $\alpha$ values. The dashed line 
    marks $\alpha = 1$.}
    \label{fig:app_gsm_alpha}
\end{figure}

Figure~\ref{fig:app_gsm_alpha} plots mean correction rate vs.\ $\alpha$ 
for each student architecture on GSM8K. The contrast with the TruthfulQA 
profiles (Figure~\ref{fig:alpha}) is stark: all four architectures collapse 
to near-zero correction rate for $\alpha \geq 0.8$, with no recovery at 
higher extrapolation values. This uniform collapse across architectures 
indicates that mathematical representations are substantially more fragile 
than verbal ones, tolerating only the most conservative interventions before 
coherent generation fails entirely.

\section{Alpha Sensitivity: TruthfulQA vs.\ GSM8K}
\label{app:alpha_comparison}

\begin{figure}[h]
    \centering
    \includegraphics[width=\linewidth]{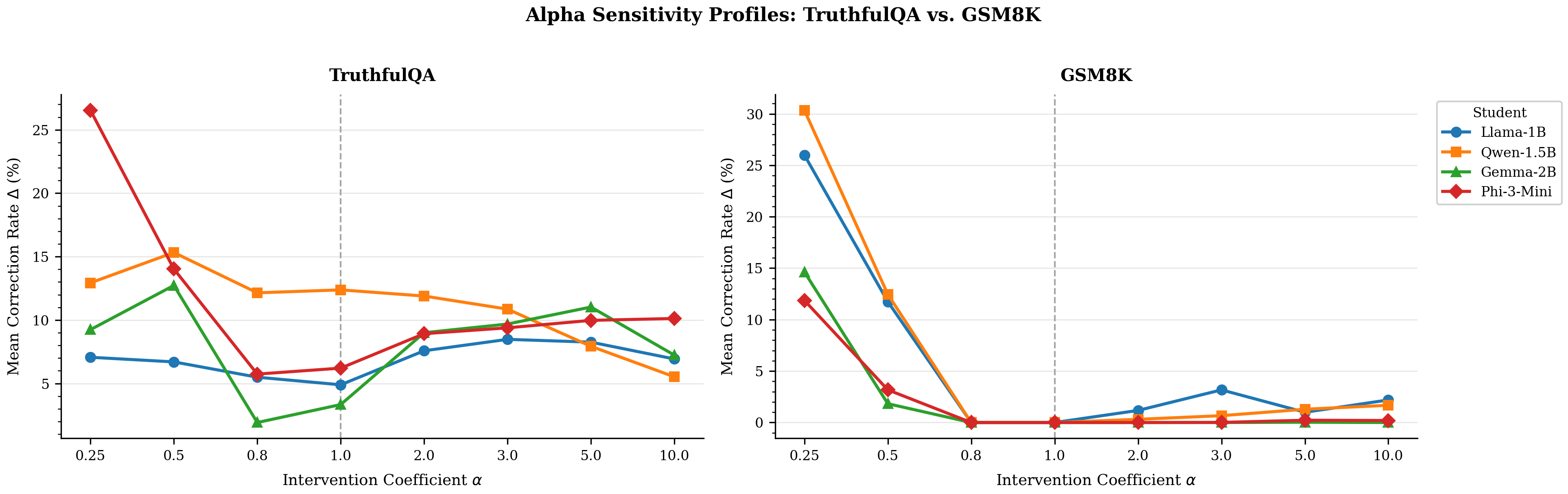}
    \caption{Alpha sensitivity profiles for TruthfulQA (left) and GSM8K 
    (right), one line per student architecture averaged across all teachers 
    and layer combinations. The dashed vertical line marks $\alpha = 1$. 
    TruthfulQA profiles are architecturally heterogeneous with partial 
    recovery at higher $\alpha$; GSM8K profiles converge uniformly on 
    $\alpha = 0.25$ with complete collapse beyond $\alpha = 0.5$ across 
    all architectures.}
    \label{fig:app_alpha_comparison}
\end{figure}

Figure~\ref{fig:app_alpha_comparison} presents TruthfulQA and GSM8K alpha 
sensitivity profiles side by side, making the domain contrast directly 
legible. On TruthfulQA, student architectures exhibit distinct and 
heterogeneous profiles with partial recovery at moderate extrapolation 
values. On GSM8K, all four architectures converge on a single behavioral 
pattern: peak performance at $\alpha = 0.25$ followed by uniform collapse. 
The architectural heterogeneity visible on the left panel disappears 
entirely on the right, suggesting that the fragility of mathematical 
representations under representational displacement is a domain-level 
property rather than an architecture-level one.

\section{GSM8K Layer Interaction Heatmap}
\label{app:gsm_layers}

\begin{figure}[h]
    \centering
    \includegraphics[width=0.78\linewidth]{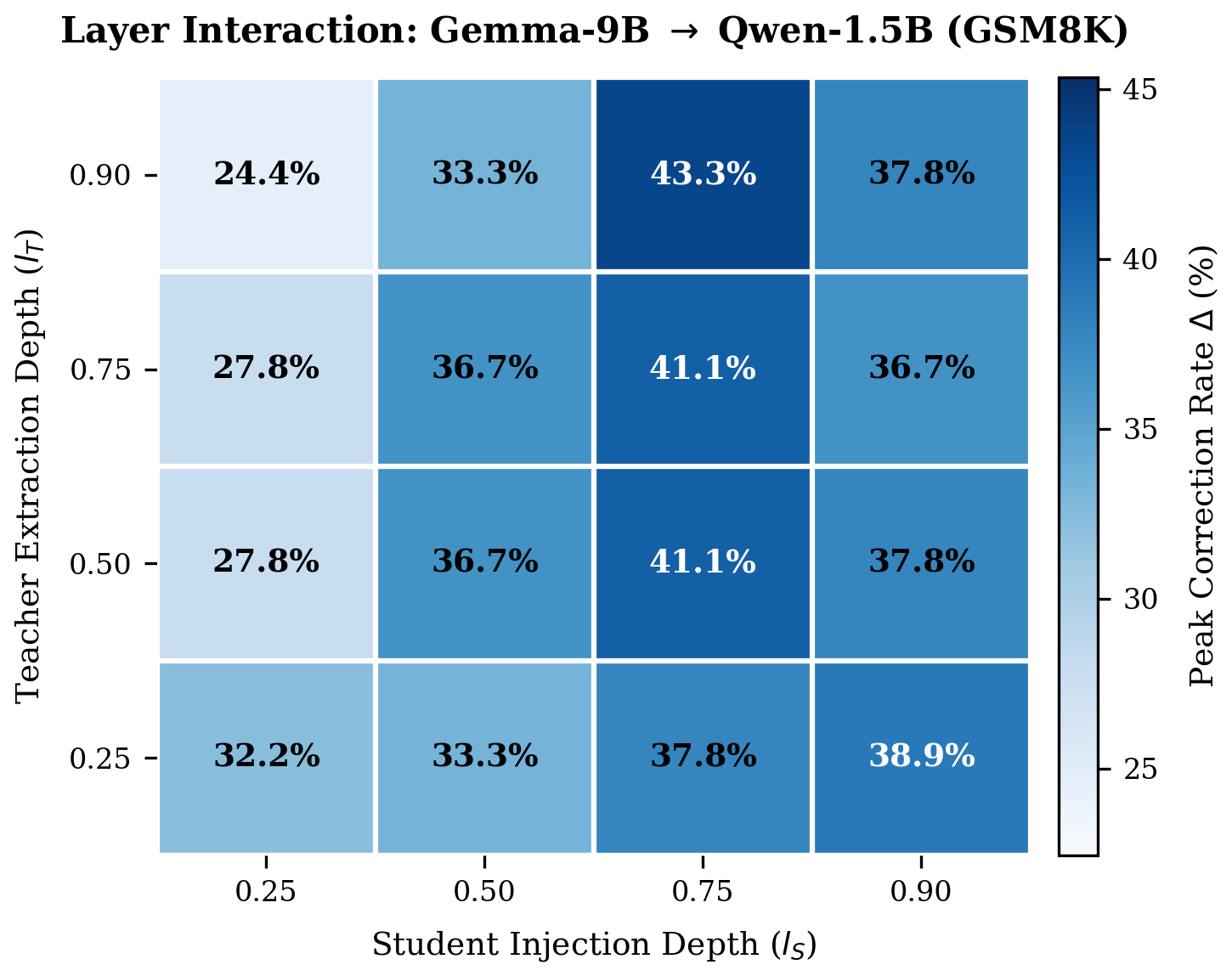}
    \caption{Peak correction rate $\Delta$ across all 16 $l_T \times l_S$ 
    combinations for Gemma-9B $\rightarrow$ Qwen-1.5B on GSM8K. Rows are 
    ordered with deepest teacher extraction at the top. The gradient with 
    teacher depth is less monotonic than the TruthfulQA counterpart, with 
    lower-depth teacher extraction performing more competitively.}
    \label{fig:app_gsm_layers}
\end{figure}

Figure~\ref{fig:app_gsm_layers} presents the layer interaction heatmap for 
Gemma-9B $\rightarrow$ Qwen-1.5B, the highest-performing pairing on GSM8K 
(peak $\Delta = 43.3\%$). Compared to the TruthfulQA counterpart 
(Figure~\ref{fig:layers}), the monotonic improvement with teacher depth is 
less pronounced, and lower teacher extraction depths perform more 
competitively. This suggests that mathematical reasoning representations 
reach a usable state of resolution at shallower teacher layers than verbal 
reasoning representations, consistent with the simpler syntactic structure 
of arithmetic problems relative to open-domain factual questions.

\section{Layer Dissociation: Full Depth Sweep}
\label{app:layer_dissociation}

\begin{figure}[h]
    \centering
    \includegraphics[width=\linewidth]{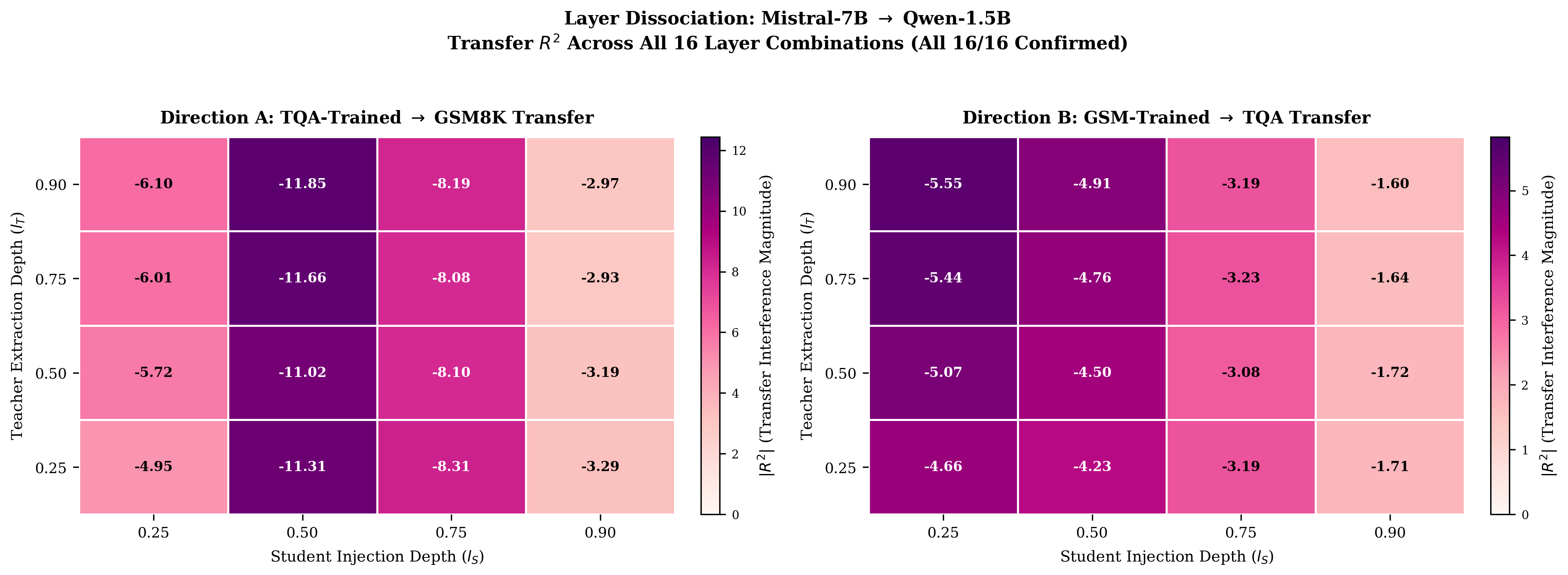}
    \caption{Transfer $R^2$ across all 16 $l_T \times l_S$ combinations 
    for Mistral-7B $\rightarrow$ Qwen-1.5B. Left panel: TruthfulQA-trained 
    mapper evaluated on GSM8K. Right panel: GSM8K-trained mapper evaluated 
    on TruthfulQA. Color intensity reflects absolute interference magnitude; 
    all values are negative. Mid-network injection ($l_S = 0.50$) produces 
    the strongest interference in both directions; late injection 
    ($l_S = 0.90$) produces the weakest. Dissociation confirmed 16/16.}
    \label{fig:app_layer_dissociation}
\end{figure}

Figure~\ref{fig:app_layer_dissociation} presents transfer $R^2$ values 
across all 16 teacher extraction depth $\times$ student injection depth 
combinations for the Mistral-7B $\rightarrow$ Qwen-1.5B pair. Color 
intensity reflects the absolute magnitude of transfer interference --- 
darker cells indicate more catastrophic geometric mismatch. Dissociation 
is confirmed in all 16/16 combinations in both transfer directions, 
establishing that domain orthogonality is not a depth-specific artifact.

A systematic pattern is visible across both panels: student injection 
depth ($l_S$) modulates interference magnitude more strongly than teacher 
extraction depth ($l_T$). Mid-network injection ($l_S = 0.50$) consistently 
produces the strongest interference (TQA$\rightarrow$GSM range: $-11.0$ 
to $-11.9$), while late injection ($l_S = 0.90$) produces the weakest 
(range: $-2.9$ to $-3.3$). Teacher extraction depth has a comparatively 
weaker and less consistent effect. This asymmetry implicates the 
mid-network student layers as the primary site of domain-specific semantic 
construction, consistent with the hierarchical processing account in 
Section~\ref{sec:temporal}.

\section{Implementation and Reproducibility Details}
\label{app:implementation}

\paragraph{Hardware.}
All experiments were conducted on a single NVIDIA A100 80GB GPU. Teacher 
and student models were loaded sequentially to manage GPU memory: the 
teacher was loaded, all activation vectors were extracted and cached to 
CPU, and the teacher was then unloaded before the student was loaded. The 
Llama-70B teacher was the only model requiring quantization, loaded in 
4-bit NF4 format using \texttt{BitsAndBytesConfig} with 
\texttt{bnb\_4bit\_compute\_dtype=}\\
\texttt{torch.float16} and 
\texttt{bnb\_4bit\_use\_double\_quant=True}. All other models were loaded 
in \texttt{bfloat16} precision.

\paragraph{Inference Configuration.}
All forward passes used left-padded tokenization 
(\texttt{padding\_side=`left'}) with \texttt{pad\_token} set to 
\texttt{eos\_token} where no dedicated pad token existed. Activation 
vectors were extracted as the final-token hidden state at each candidate 
layer using \texttt{output\_hidden\_states=True}. Batch size was 32 for 
all extraction and intervention passes. Generation used greedy decoding 
(\texttt{do\_sample=False}, \texttt{max\_new\_tokens=50} for TruthfulQA, 
\texttt{max\_new\_tokens=100} for GSM8K).

\paragraph{Intervention Hook.}
The residual stream intervention was implemented as a PyTorch forward hook 
registered on \texttt{model.model.layers[layer\_idx]}. The hook intercepts 
the layer output, extracts the final token position, rescales the 
pre-computed projected vector to match the current residual stream norm, 
applies the weighted combination from Equation~\ref{eq:intervention_full}, 
and returns the modified hidden state. The hook is registered immediately 
before each batched generation call and removed immediately after, 
ensuring no cross-contamination between intervention and baseline passes.

\paragraph{Projection Matrix Estimation.}
Ridge and Lasso regressors were implemented using \texttt{scikit-learn}'s 
\texttt{Ridge(alpha=0.1)} and \texttt{Lasso(alpha=0.0001, max\_iter=5000)} 
with default \texttt{fit\_intercept=True}. The 70/30 train/test split 
used \texttt{sklearn.model\_selection.}\\ \texttt{train\_test\_split} with 
\texttt{random\_state=42}. Permutation controls used 
\texttt{numpy.random.permutation} on the training split target matrix 
prior to fitting, with the same fixed seed.

\paragraph{Evaluation.}
Generated text was decoded with \texttt{skip\_special\_tokens=True}, 
split on the \texttt{"Answer:"} delimiter to isolate the model's response, 
and lowercased and stripped before substring matching. Both the 
\texttt{best\_answer} and all entries in \texttt{correct\_answers} were 
lowercased prior to matching. A response was counted as correct if any 
reference string appeared as a substring of the generated text.

\end{document}